# Accurate Automatic Segmentation of Amygdala Subnuclei and Modeling of Uncertainty via Bayesian Fully Convolutional Neural Network

Yilin Liu[1], Gengyan Zhao[2], Brendon M. Nacewicz[3], Nagesh Adluru[1],
Gregory R. Kirk[1], Peter A Ferrazzano[1,4], Martin Styner[5,6], Andrew L Alexander[1,2,3]*

**Affiliations:**
1 Waisman Brain Imaging Laboratory, University of Wisconsin – Madison,
  Madison, Wisconsin, United States
2 Department of Medical Physics, University of Wisconsin – Madison,
  Madison, Wisconsin, United States
3 Department of Psychiatry, University of Wisconsin – Madison,
  Madison, Wisconsin, United States
4 Department of Pediatrics, University of Wisconsin – Madison,
  Madison, Wisconsin, United States
5 Department of Psychiatry, University of North Carolina – Chapel Hill,
  Chapel Hill, North Carolina, United States
6 Department of Computer Science, University of North Carolina – Chapel Hill,
  Chapel Hill, North Carolina, United States

* **Corresponding Author:** Andrew L. Alexander, PhD., Waisman Brain Imaging Laboratory, Waisman Center, 1500 Highland Ave, Madison, WI 53705. E-mail: alalexander2@wisc.edu

**Abstract**

Recent advances in deep learning have improved the segmentation accuracy of subcortical brain structures, which would be useful in neuroimaging studies of many neurological disorders. However, most of the previous deep learning work does not investigate the specific difficulties that exist in segmenting extremely small but important brain regions such as the amygdala and its subregions. To tackle this challenging task, a novel 3D Bayesian fully convolutional neural network was developed to apply a dilated dual-pathway approach that retains fine details and utilizes both local and more global contextual information to automatically segment the amygdala and its subregions at high precision. The proposed method provides insights on network design and sampling strategy that target segmentations of small 3D structures. In particular, this study confirms that a large context, enabled by a large field of view, is beneficial for segmenting small objects; furthermore, precise contextual information enabled by dilated convolutions allows for better boundary localization, which is critical for examining the morphology of the structure. In addition, it is demonstrated that the uncertainty information estimated from our network may be leveraged to identify atypicality in data. In an evaluation of 14 T1-weighted images with expert defined amygdalae and subregions the method demonstrated a Dice overlap coefficient of 0.910 for the amygdala and an average 0.804 across its subregions. Our method was compared with two state-of-the-art deep learning models and a traditional multi-atlas approach, and exhibited excellent performance as measured both by Dice overlap as well as average symmetric surface distance. To the best of our knowledge, this work is the first deep learning-based approach that targets the subregions of the amygdala. The code of our method will be made publicly available to the neuroimaging community.

## 1. Introduction

The amygdala is a key regulator of emotional arousal and is thought to regulate generalization or habituation of fear responses in normal and abnormal development (Adolphs et al. 2005; Öhman, 2005; Knight et al., 2005). Animal models have been used to differentiate subregions of the amygdala, identifying structural bases of fear generalization in basal and lateral nuclei distinct from output projections from centromedial regions (Hrybouski et al. 2016, LeDoux,2007; Kwapis et al. 2017), and reliable quantification of these substructures is urgently needed. An accurate segmentation critically influences the quantitative analysis of the amygdalae and characterization of the associated neuropsychiatric diseases. However, as a deep heterogenous cluster of subregions, surrounded by vasculature and sources of MRI field inhomogeneities, it remains an extremely difficult region to quantify. Compared with conventional automated software (Freesurfer, FSL), hand drawn amygdala boundaries can better capture cumulative contributions of biological and environmental stress, including autistic social impairment, physical abuse, institutional neglect and poverty (Nacewicz et al., 2006; Hanson et al., 2015). However, manual segmentation is often time-consuming and is prone to biases (Maltbie et al., 2012), highlighting the need for highly accurate automated segmentation methods.

Recently, convolutional neural networks (CNNs) have shown to outperform traditional segmentation methods in various computer vision tasks and have been investigated in many medical applications with extremely promising results (Roneberger et al., 2015; Lv et al.,2017; Mahapatra et al.,2018; Nie et al.,2017). However, most of the existing deep learning-based approaches either focus only on segmenting large subcortical structures (such as thalamus, putamen, caudate, pallium) (Dolz et al., 2017; Shakeri et al., 2016), or do not obtain optimal results on small but important structures such as the amygdala, let alone the subregions. Indeed, segmenting extremely small structures using CNN methods inherently poses several challenges. First, smaller structures results in smaller targets in size for training, making the dataset highly imbalanced. This often leads to bias towards the prediction of background for a cost-insensitive classifier whose goal is to maximize the overall accuracy (or minimize the overall error rate) regardless of classes. Second, incorporating contextual information, such as a structure's surrounding, and retaining fine details is often a tradeoff within a CNN, yet it is important to optimize both of these aspects to adequately recognize small objects (Hu & Ramannan, 2016; Mottaghi et al., 2014). Therefore, in order to segment small subcortical structures with high accuracy and robustness, it is necessary to take these difficulties into account in designing CNN architectures.

*1.1 Related work*

A large number of segmentation methods for the amygdala have been proposed, which may be classified into atlas-based and learning-based categories. A traditional representative is the single atlas-based method in which the segmentation of a target image is estimated by aligning it with a manually labeled template through registration (Wu et al., 2007). However, mismatches may easily occur when the target image is sufficiently different from a single atlas. Challenges may also arise from differences in MRI instrumentation or protocols as well as inherent topological differences across individuals. To account for such variability, multi-atlas methods which use multiple labeled atlases have been developed (Heckemann et al., 2006; Wang et al., 2012; Rohlfing et al.,2004). Several studies have reported that multi-atlas methods yield high agreement with the manual ground truth (Hanson et al., 2012; Babalola et al., 2009; Leung et al., 2005). Nevertheless, there is a considerable computational cost since all of the atlases need to be registered to each target image case using non-linear deformable transformations (Hanson et al., 2012). Additionally, the segmentation quality in multi-atlas approaches highly depends on the selection of the atlases and the fusion algorithm (Rohlfing et al. 2004; Aljabar et al., 2009). Other automatic atlas-based segmentation packages are FreeSurfer and FIRST from FSL. FreeSurfer segments voxels based on probabilistic information estimated from a large expert-labeled training dataset (Fischl et al., 2002), and FIRST uses Bayesian shape and appearance models (Zhang et al., 2001). It has been reported that amygdala segmentation with FreeSurfer correlates more highly with manual tracing than FIRST, while FIRST is better in representing the amygdala shape than FreeSurfer, but overall their segmentation performances remain suboptimal (Morey et al., 2009; Schoemaker et al., 2016) due to insensitivity to biologically-relevant variance (Hanson et al., 2015).

Recently, convolutional neural networks (CNN) have brought tremendous improvements in various computer vision tasks such as image classification and semantic segmentation (Krizhevskey et al., 2012; Simonyan et al., 2014; He et al., 2015). Unlike traditional machine learning, CNNs can autonomously learn representations of data with increasing levels of abstraction via multiple convolutional layers without any feature engineering. In CNNs, weights are shared and locally connected among convolutional layers, which significantly reduces the number of parameters compared with fully connected layers, making CNNs especially suitable for imaging tasks. Naturally, CNNs have been gradually becoming the tool of choice for medical imaging tasks. In image segmentation, a classification network was previously used under a sliding window scheme to predict the class probability of the center pixels of over-lapping patches (Ciresan et al., 2013). Since such a classification makes predictions for a single pixel at a time, this approach suffers from redundant computations and does not benefit from correlations across pixels. Long et al. (2014) first proposed fully convolutional neural networks (FCNN) in which the fully connected layers are replaced with 1x1 convolution so that the network consists of convolutional layers only. This strategy allows dense predictions for multiple pixels in a single forward pass, and eliminates the limitation posed by fully connected layers on the size of the inputs to accept input images with arbitrary size. FCNN therefore serves as an effective general purpose engine for tasks of semantic segmentation.

A popular FCNN in medical imaging applications is "U-Net", proposed by Ronneberger et al. (2015). It consists of an encoder-decoder architecture that can be laid out as a U shape on a page. The encoder part compresses the input images into lower-resolution feature maps via downsampling or pooling layers, and the decoder part aims to recover the full-resolution label map from these feature maps for pixel-to-pixel semantic classification. Similar architectures are seen in 3D U-Net (Cicek et al., 2016), V-Net (Milletari et al., 2016) and SegNet (Badrinarayanan et al., 2016). These networks have similar encoders - a VGG-like (Simonyan et al., 2014) architecture is typically adopted, while vary in their decoders. Multiple up-sampling strategies have been proposed for decoders, including deconvolution (Noh et al., 2015), bilinear upsampling and unpooling (Chen et al., 2015; Badrinarayanan et al., 2016). However, such design could pose a few problems when segmenting structures with small spatial extent. First, although consecutive strided convolutions or pooling operations employed in these networks enable a large receptive field, fine details may be lost and are difficult to recover via simple non-learnable upsampling strategies or skip connections. For example, if a network has a downsample rate of 1/8 (as it employs three max-pooling layers with 2x2 filters with stride 2), an object with less than 8 voxels (such as the amygdala's subregions) in each dimension is not likely to be recovered later. Second, since down-sampling operations typically lead to great dimension reduction, the input images of these networks need to be large enough so as to preserve sufficient dimension after the compression of the encoder, for being further processed by the decoder. But larger image patches are more likely to be dominated by background voxels compared with smaller ones, leading to severe class imbalance problem. This makes the predictions more favorable to the background, which is particularly of concern for small objects. Although a weighted cross entropy loss function has been suggested to alleviate this problem (Ronneberger et al., 2015; Cicek et al., 2016), choosing a proper weight map for all the classes is non-trivial. Another solution could be the Dice loss function (Milletari et al., 2016) which avoids tuning any extra hyperparameter and weighs false negatives and false positives equally. Hence, although these networks have plenty of success in segmentation tasks of large structures such as brain extraction (Zhao et al., 2018), lung (Chon et al., 2017) and breast segmentation (Dalmış et al., 2017), specific strategies for small structures are necessary.

Compared with large structures, smaller ones provide less information for training due to the smaller volumes, which makes the learning of discriminative features more challenging. Hu and Ramanan (2016) suggested that modeling context is particularly helpful for CNNs to recognize small objects, based on a key observation that humans can only accurately classify small faces with evidence beyond the object itself. In general, context can provide knowledge of a structure with respect to its surroundings and disambiguate objects with similar local visual appearances. Thus, incorporating context can critically improve recognition accuracy (Galleguillos & Belongie, 2010). In medical imaging, many studies have explored the idea of using input patches with various sizes for modeling multi-scale contextual information (Moeskops et al., 2016; Brebisson & Montana, 2015; Ghafoorian et al., 2017; Mehta et al., 2017; Kamnitsas et al., 2015). Most of these networks are organized in a multi-branch manner, where each branch independently processes patches of a certain type. In order to reduce the computational complexity brought by larger patches, these methods typically involve down-sampling as the first step, thereby providing the networks coarse "cues" about the surroundings of the regions of interest. In other patch-based CNN approaches, explicit spatial features obtained from a structural probabilistic atlas are combined with CNN features to provide additional spatial information (Kushibar et al., 2018). Segmenting small structures with high accuracy is therefore reduced to the problem of finding the optimal trade-off between capturing sufficiently large context and retaining fine details, while alleviating the imbalanced class issue.

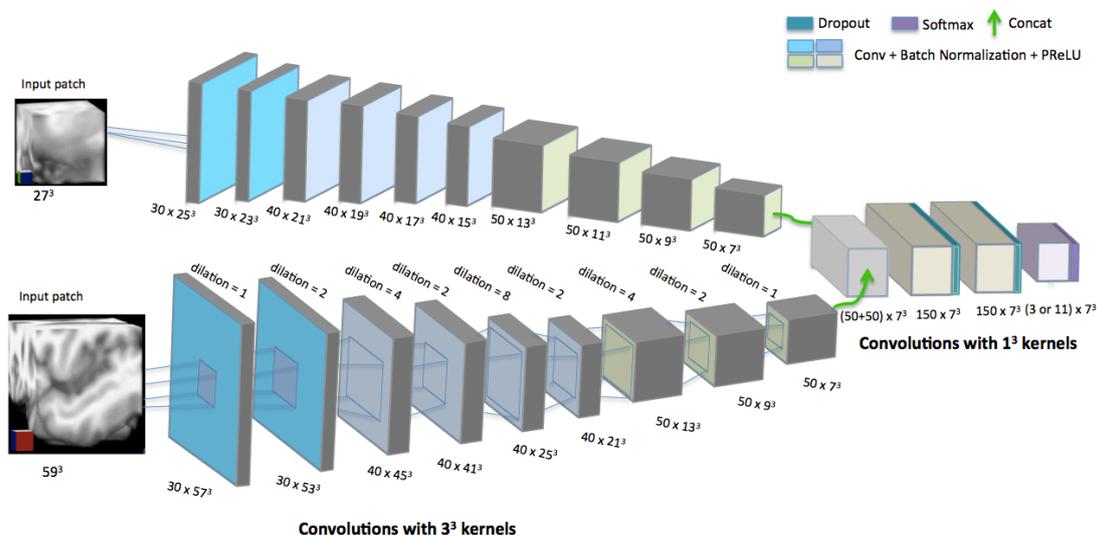

Fig.1. Overview of the proposed CNN structure. The kernels used throughout the upper pathway are of size $3^3$, resulting in a receptive field of size $21^3$. The kernels passed on the first layer of the bottom pathway are also of size $3^3$, but the actual sizes are enlarged as the dilation rate increases and thus results in a receptive field of size $53^3$. Number of feature maps and their size denoted as *Number x Size.*

*1.2 Overview of our approach*

In light of the limitations of previous works, we propose a 3D Bayesian FCNN that is designed for segmenting small sub-cortical structures with high accuracy. Specifically, our network consists of a decoupled dilated dual-pathway, which can exploit both local and global contextual information and can maintain fine details without increasing computational complexity. In addition, Monte Carlo (MC) dropout is integrated in the fully automatic pipeline, in order to provide uncertainty information. In this way, we are able to estimate the confidence of the model about unseen data in the absence of ground truth, which can be informative for identifying atypical data in our experiments. The effects of data augmentation are also investigated. The performance of each single pathway is analyzed and compared with the final dual-path model, in order to illustrate the improved efficacy of our network design. We compare our method with two state-of-the-art deep learning approaches and a multi-atlas method.

## 2. Data and Methods

*2.1. Dataset*

T1-weighted MRI data from 14 subjects (age mean (standard deviation) 28.9y (6.5y); range 18.5-43.4y) were collected on a GE ME750 3.0 T MRI scanner with the product 8-channel head coil. All participants provided written consent or assent as part of a procedure approved by the Human Subjects Institutional Review Board of the University of Wisconsin School of Medicine and Public Health. Dimensions of each volume is 171 x 236 x 191. Four separate imaging sessions were performed on each subject and co-registered and averaged via a series of half-transforms described in a prior study (Nacewicz et al., 2012). The pre-processing pipeline involves skull-stripping, bias-field correction and volume-wise intensity normalization. Images are resampled to 171 x 236 x 191 bricks of 1mm isotropic voxels from an original resolution of 240x240x124mm (0.9375x0.9375x1mm) in a landmark-based AC-PC alignment, followed by rotation in sagittal plane to the "pathological plane" to match post-mortem atlases. The left and right amygdala and four subnuclear groups on each side - lateral, basal, cortico-superficial (ofactory) and centromedial subregions - were manually labeled by an amygdala anatomy expert (BN). Overall, the right and the left amygdalae jointly account for about 0.05% of the whole brain volume of a single subject.

*2.2. Dilated Convolution*

Recently, Yu and Koltun (2016) proposed dilated convolution as a method that can expand the receptive field without downsampling the images. Specifically, convolutional kernels are upsampled by inserting "holes" (zeros) between weights. This was originally developed for wavelet decomposition, known as "*algorithme atrous*" (Holscheider et al., 1990). In a CNN with dilated convolutions, multi-scale contextual information can be systematically aggregated without sacrificing resolution. Also, this does not increase the computational complexity since only non-zero values are taken into account during convolutions. The receptive field in layer $l$ can be defined as:

$$RF_{K_l} = k + (k-1) \times (D_k - 1),$$

where K denotes the kernel size and D denotes the dilation rate. The gap between elements in a kernel is $D_k - 1$. In standard convolution, $D_k$ is 1. However, the use of dilated convolutions inherently causes *gridding* artifacts. Since units within kernels are sparsely connected (i.e., zeros are padded between two units in a kernel), locations with checkerboard patterns are covered, leading to loss of neighboring information. This problem becomes more severe when the dilation rate is aggressively increased. To alleviate this problem, Yu et al. (2017) adopted a strategy that dilation rates are first increased and then progressively lowered as the depth of the network increases. Similar in spirit, we employed oscillating dilation convolutions, i.e., convolutions with higher dilation rates are alternated with ones with lower dilation rates.

*2.3 Bayesian convolutional nerual network*

Bayesian inference as a principled technique to estimate model uncertainty had rarely been used in CNNs due to prohibitive computational cost. Recently, Gal and Ghahramani (2015) showed that dropout training can be casted as approximate Bernoulli variational inference to allow an efficient approximation of the model's posterior distribution without additional parameters. Namely, a Bayesian CNN can simply be implemented by performing dropout after convolution layers, which is equivalent to placing a Bernoulli distribution over the weights, and training with dropout is in effect the process of minimizing the Kullback-Leibler divergence between the approximating distribution and the posterior distribution over the weights. At testing time, by retrieving N stochastic outputs from the network with dropout, the posterior distribution can then be approximated, referred to as Monte Carlo (MC) Dropout (Gal & Ghahramani, 2016). The mean and variance of these samples can be interpreted as the segmentation output and uncertainty estimate, respectively (Gal &

Ghahramani, 2015). Compared with the standard weight averaging technique in which dropout is turned off during testing and the trained weights are scaled down by the dropout rate *p*, Monte Carlo sampling has been shown to lead to better accuracy in various recognition tasks (Kendall et al.,2016; Gal & Ghahramnani, 2015; Gengyan Zhao et al., 2018). Therefore, in this study, we employed the Monte Carlo sampling during testing.

*2.4 The Full Architecture Framework*

In our network design, we adopted the strategies to alleviate class imbalance, retain fine details and incorporate high-resolution contextual information. Inspired by DeepMedic (Kamnitsas et al., 2017), we proposed a variant of it for our segmentation tasks. Specifically, our network consists of decoupled dual dilated convolutional pathways (Fig. 1), with one pathway specializing in learning local detailed features and the other one incorporating large contextual information. The key distinction from DeepMedic is that the second pathway of our model does not operate on a down-sampled version of the image that has lower resolution; rather, it directly learns from the original image through dilated convolutions for precise location prediction. Additionally, the feature maps from the two pathways are combined directly without non-learned up-sampling schemes (simple repetition) which may introduce inaccurate information. The necessity of this design will be discussed later.

To retain fine details, pooling layers are *not* employed in this network. Furthermore, unary strides are used in all convolution operations throughout the network. All convolutions are applied without padding. Each convolutional layer is followed by a batch normalization layer (Ioffe and Szegedy, 2015) for speeding up the training, and is then applied with PReLU non-linearity (He et al., 2015), which can be defined as follows:

$$f(y_i) = \begin{cases} y_i, & if\ y_i > 0 \\ a_i y_i, & if\ y_i \leq 0 \end{cases}$$

where the slope parameter $a$ can be adaptively learned during training, making the activation more specialized for the task with negligible computational cost.

The number of kernels passed on each 3 x 3 x 3 convolutional layer in the first and the second pathways are as follows: 30, 30, 40, 40, 40, 40, 50, 50, 50, 50, and 30, 30, 40, 40, 40, 40, 50, 50, 50, respectively. Note that the lengths of the two pathways are different. The first pathway with a small receptive field of 21 x 21 x 21 that can enclose the whole amygdala without excessive background voxels allows for better texture analysis and alleviates the class imbalance problem. The dilation rate was set to 1,2,4,2,8,2,4,2,1 in the second pathway, resulting in a receptive field of 53 x 53 x 53. Feature maps (FM) of the first and second pathways are combined before the two fully connected layers through direct element-wise concatenation. To allow efficient dense predictions, the fully connected layers are converted to 1 x 1 x 1 convolutions, following the strategy of (Long et al., 2014). The final classification layer reduces the number of channels to the number of labels which is either 3 or 11 (including the background), depending on the task. The softmax function attached to the classification layer yields the segmentation result in the form of a spatial probability map.

To turn a standard CNN into a Bayesian CNN for capturing the uncertainty, we extend our network with dropout layers after the PReLU non-linearity, which not only allows for efficient Monte Carlo sampling but also reduces over-fitting. As applying dropout in all layers potentially generates too strong of a regularizing effect and thus results in suboptimal training (Kendall, 2015), we only apply dropout in the last two 1x1x1 convolutional layers (before the classification layer), which form higher-level features and contain the most parameters (Fig. 1).

*2.5. Training*

We implemented our network (https://github.com/YilinLiu97/AmygNet_subregions) and conducted the network training on the NiftyNet framework (Gibson et al., 2017). We also evaluated DeepMedic (Kamnitsas et al., 2017) and LiviaNET (Dolz et al., 2017) on the same dataset. DeepMedic was also trained on the NiftyNet framework, and LiviaNET was trained using the provided source code (https://github.com/josedolz/LiviaNET). The networks were trained respectively for segmenting (a) the full left and right amygdala and (b) their subregions.

**Data augmentation:** To alleviate the problem of low sample size, we performed online data augmentation. Specifically, the training patches were randomly applied with [-10°, 10°] rotation operations and/or were scaled down/up by a factor in the range [0.8, 1.2]. The number of patches were about 27500 and 33000 patches for the full amygdala and subnuclear segmentation task, respectively. The segmentation results before and after data augmentation were compared.

**Sampling:** Since MRI brain volumes are generally large, there were computation and GPU memory constraints. Thus, volumes were split into smaller segments for training and dense inference. The inputs of the two pathways were centered at the same point on the image. In each iteration, 11 patches of size 59 x 59 x 59 were sampled from the whole brain, within which small patches of size 27 x 27 x 27 were cropped and fed into the local context pathway. The entire 59 x 59 x 59 patches were then fed into the global context pathway. During inference, 105 x 105 x 105 patches were used. We observe that a balanced extraction of samples is a critical factor for successful experiments in our case. Namely, it's crucial to ensure that each label has same probability of occurrence.

**Hyperparameter selection:** We employed the categorical cross entropy as the cost function and applied optimization via the Adam method with a fixed learning rate of 0.001. Training was performed in batches of 11 image patches during each iteration. Weights in each layer were initially drawn from a zero-based Gaussian distribution with standard deviation of $\sqrt{2/n_i}$, where $n$ denotes the number of units in a kernel of the layer $l$ (He et al., 2015). Bias were initialized at zero. Dropout rate was set to 0.3 at both the training and testing phases. The training of the amygdala, and the subnuclear segmentation tasks were stopped after 2500 iterations and 3000 iterations respectively, as they were found to be sufficient for convergence. Same parameters were used for training DeepMedic. Default parameters were used for training LiviaNET (momentum of 0.6; initial learning rate of 0.001, decayed every 2 epochs by a factor of 2; 20 subepochs per epoch; 1000 samples in each subepoch). For all evaluated methods, a leave-one-out cross validation was performed for both tasks. In each training fold, 12 subjects were used for training and the remaining subjects were used for validation and testing, respectively. The model parameters in the epoch that results in best performance (i.e., highest average dice score) on the validation subject were used to segment the test subject.

*2.6. Evaluation metrics*

The pair-wise similarity and discrepancy of our automatic (A) and manual segmentation (M) were evaluated using the commonly employed Dice Similarity Coefficient (DSC) (Dice, 1945). Dice values range from zero to 1, where 1 indicates 100% overlap with the ground truth, and 0 indicates no overlap. However, volumetric overlap measures are not sensitive to the contour of the segmentation output, while the latter is important in many medical applications such as disease diagnosis and treatment planning, as is also the case for the amygdala (Shenton et al., 2002; Tang et al., 2015; Yoon et al., 2016). Thus, we additionally consider a distance-based metric – the average symmetric surface distance (ASSD) in our evaluation, which is defined as the average of distances between border voxels of our automatic segmentation output and those of manual segmentation output (Geremia et al., 2011):

$$ASSD = \frac{1}{|A| + |M|} \left( \sum_{a \in A} \min_{m \in M} d(a,m) + \sum_{m \in M} \min_{a \in A} d(m,a) \right),$$

Where $d(a, m)$ is the distance between point a and m. Zero value for this measure indicates perfect segmentation.

## 3. Experiments and Results

### 3.1. Comparisons with other CNN methods

We evaluated the performance of our method and two open-source deep learning models – DeepMedic (Kamnitsas et al., 2016), LiviaNET (Dolz et al., 2017) using the same leave-one-out cross validation scheme. The mean Dice coefficient and average symmetric surface distance of each structure for each method are shown in Table 1. The boxplots in Figure 1 show the median and quartiles of the DSC and ASSD for each method. Multiple pairwise Wilcoxon signed rank tests (two-sided) were used to compare the performance of these methods. Qualitative comparisons of the segmentation results obtained from DeepMedic, LiviaNET and our method are shown in Fig.2. Note that no data augmentation was performed in all these experiments for comparison purposes.

According to Table 1, our method has achieved the best mean values as well as the smallest standard deviation of DSC and ASSD in both tasks. In whole amygdala segmentation, our method is significantly better than LiviaNET in terms of both DSC ($p = 0.016$) and ASSD ($p < 10^{-5}$). Moreover, LiviaNET yields a particularly low DSC on one subject, as shown in the boxplot (Fig. 1). Compared to DeepMedic, our method also shows better performance in terms of both Dice and ASSD, though the differences are not significant ($p = 0.101$; $p = 0.165$).

On subnuclear segmentation, our method is superior to DeepMedic on all the four subnuclear regions in terms of the ASSD. In particular, the differences on the lateral and cortical-superficial subregions are significant ($p=0.036$; $p=0.006$) but not the basal nucleus and the centromedial regions ($p=0.053$; $p=0.439$). LiviaNET shows higher DSC values on lateral nucleus ($p=0.101$) and cortical-superficial subregions ($p=0.139$), while our method performs better on basal ($p=0.306$) and centromedial subregions ($p=0.891$), though all these differences are insignificant. However, as shown in the boxplot (Fig.1), LiviaNET has an outlier with extremely low DSC on every structure. The mean ASSD value of our method is significantly better than that of LiviaNET ($p < 10^{-5}$). Also, our method yields lower ASSD values on every subregion. Especially, the differences on the basal nucleus ($p < 10^{-5}$) and centromedial subregions ($p=0.008$) are significant.

|  | DeepMedic | | LiviaNET | | Our Method | |
| --- | --- | --- | --- | --- | --- | --- |
|  | DSC | ASSD/mm | DSC | ASSD/mm | DSC | ASSD/mm |
| *Amygdalae* | 0.905±0.023 | 0.360±0.400 | 0.888±0.087 | 1.677±1.012 | **0.910±0.022** | **0.256±0.368** |
| L.Amygdala | 0.904±0.025 | **0.309±0.369** | 0.874±0.121 | 1.659±1.285 | **0.910±0.027** | 0.315±0.502 |
| R.Amygdala | 0.906±0.022 | 0.411±0.435 | 0.901±0.027 | 1.696±0.689 | **0.910±0.017** | **0.197±0.149** |
| *Subregions (L,R)* | 0.768±0.074 | 1.840±2.538 | 0.778±0.121 | 1.143±0.987 | **0.781±0.069** | **0.722±0.710** |
| Lateral | 0.818±0.069 | 2.409±3.010 | **0.849±0.062** | 1.107±1.201 | 0.841±0.050 | 0.858±0.712 |
| Basal | 0.777±0.060 | 1.568±2.493 | 0.779±0.144 | 0.804±0.602 | **0.785±0.059** | **0.365±0.189** |
| Cortico-Superficial | 0.707±0.069 | 2.453±2.226 | **0.732±0.107** | 1.675±1.186 | 0.728±0.060 | 1.129±0.911 |
| Centromedial | **0.771±0.051** | 0.933±2.142 | 0.751±0.126 | 0.986±0.592 | 0.769±0.058 | **0.535±0.582** |

Table.1. Comparison of our method with DeepMedic and LiviaNET on both amygdala and subnuclear segmentation in terms of DSC and ASSD, and standard deviation. Highest DSC and ASSD values are shown in bold.

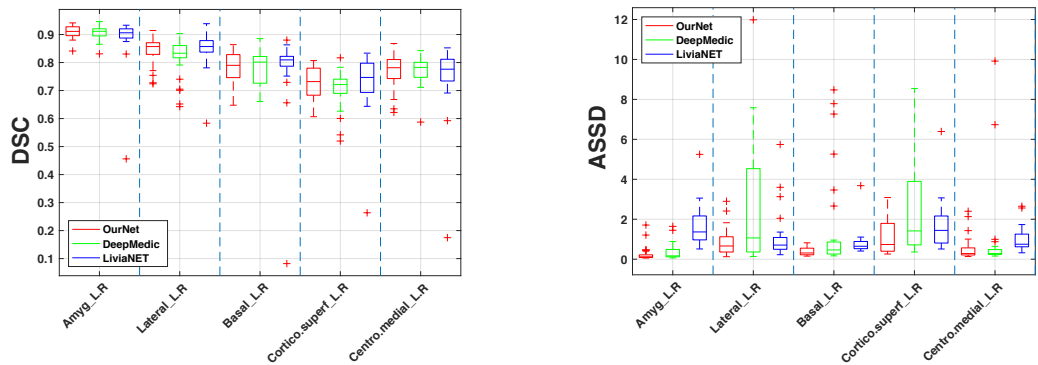

Fig. 1. Evaluation scores in boxplots from DeepMedic, LiviaNET and our method on both amygdala and subnuclear segmentation in terms of DSC and ASSD. Outliers are shown as red '+' symbols.

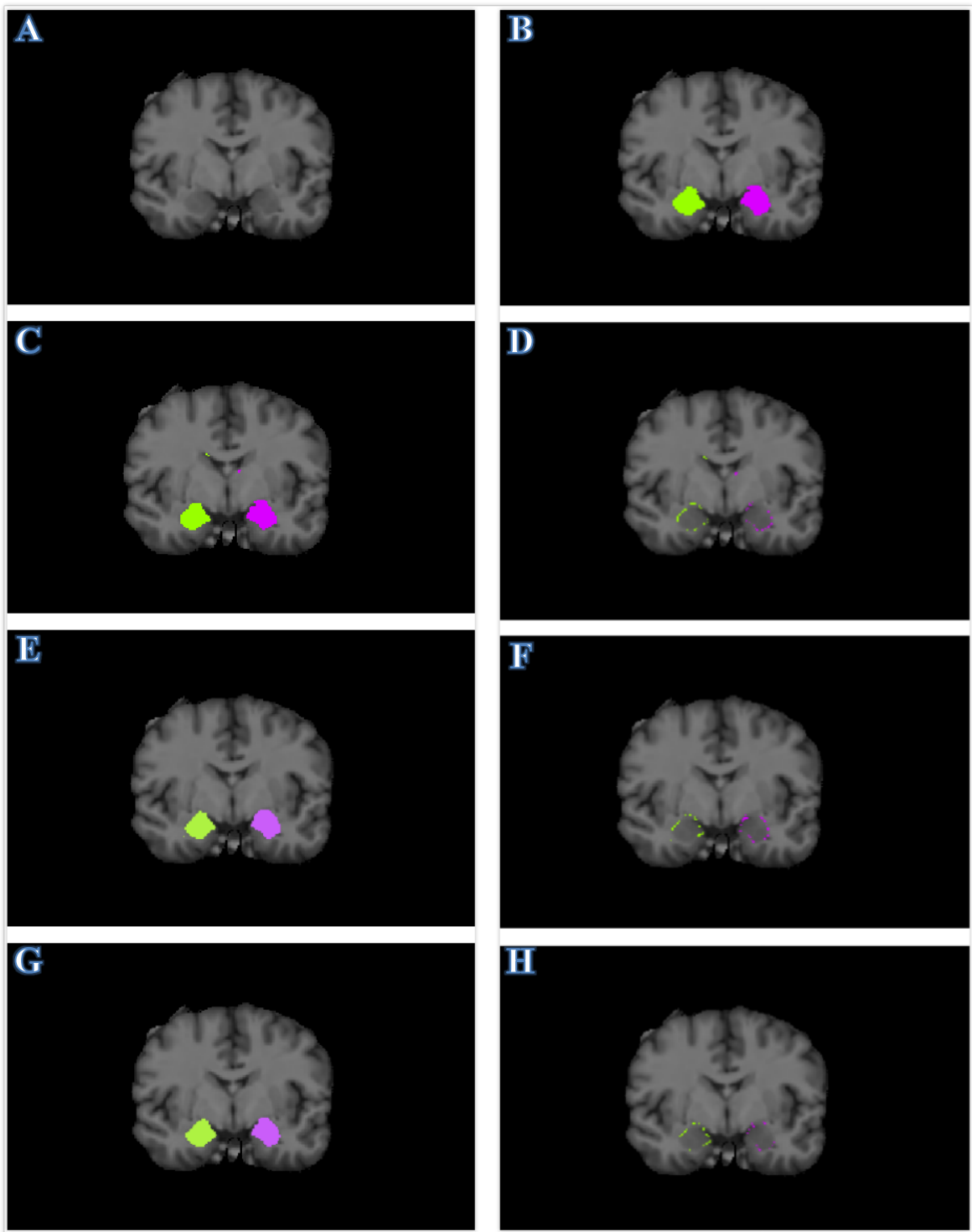

Fig. 2. Illustrations of the segmentation outputs and the misclassified voxels.
(a) Amygdala segmentation.
(A, B) T1-w image and the corresponding ground truth; (C, D) segmentation results obtained from LiviaNET and the difference from ground truth; (E, F) segmentation results obtained from DeepMedic and the difference from ground truth; (G, H) segmentation results obtained from our method and the difference from ground truth.

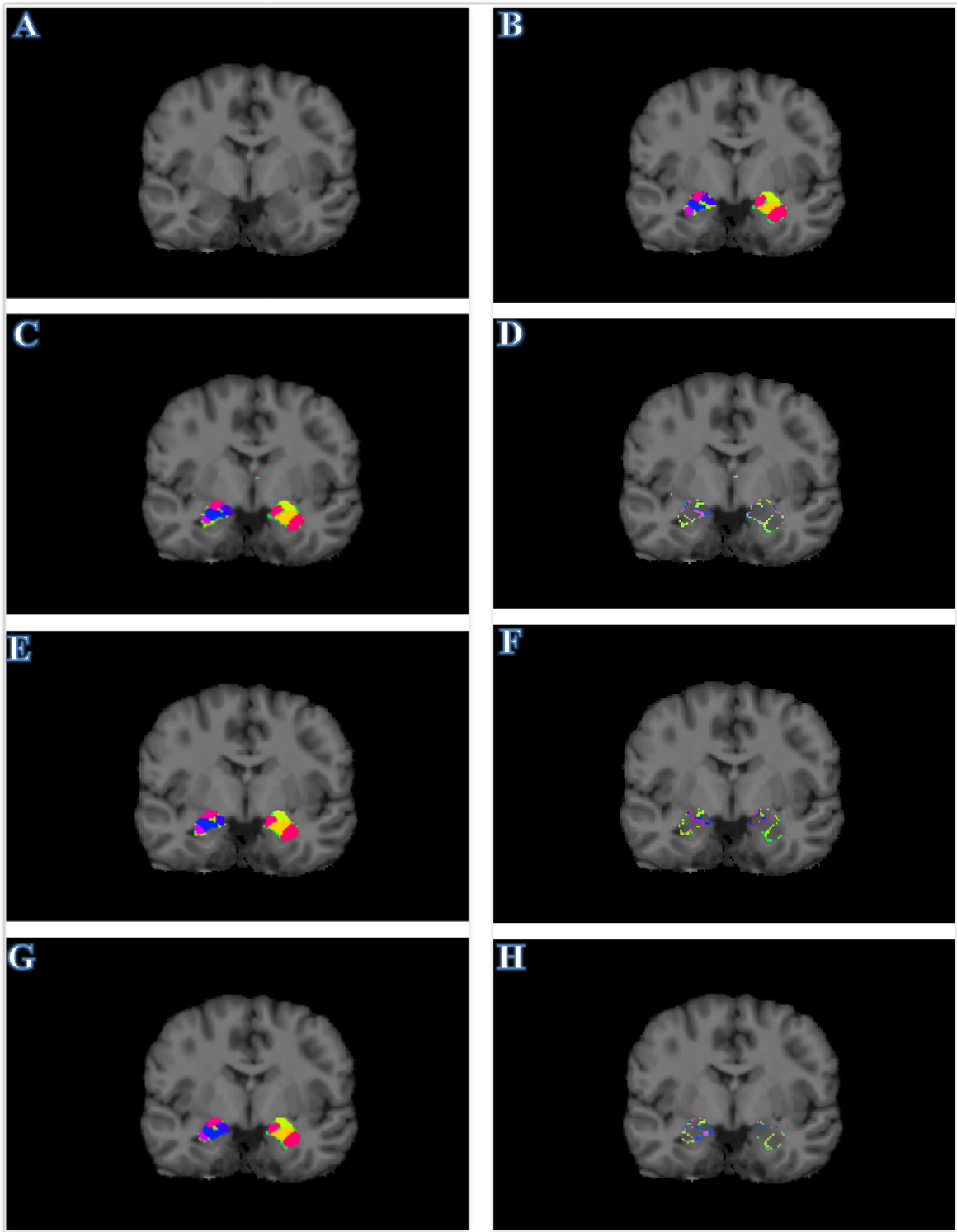

(b) Subregional segmentation.

(A, B) T1-w image and the corresponding ground truth; (C, D) segmentation results obtained from LiviaNET and the difference from ground truth; (E, F) segmentation results obtained from DeepMedic and the difference from ground truth; (G, H) segmentation results obtained from our method and the difference from ground truth.

*3.2. Effect of data augmentation*

The effects of data augmentation in both the amygdala and the subnuclear segmentation tasks were investigated and is shown in Table 3. Results show that the mean Dice Coefficients and the mean average symmetric surface distance in both tasks were improved after augmentation. In particular, data augmentation brought significant improvement to the subnuclear segmentation on both DSC and ASSD values ($p < 10^{-6}$; $p = 0.014$), while on the amygdala segmentation, the p values on DSC and ASSD values are 0.982 and 0.750, respectively.

|  | w/o. augmentation | | w. augmentation | |
| --- | --- | --- | --- | --- |
|  | DSC | ASSD/mm | DSC | ASSD/mm |
| *Amygdalae* | 0.910±0.022 | 0.256±0.368 | 0.910±0.018 | 0.223±0.242 |
| L.Amygdala | 0.910±0.027 | 0.315±0.502 | 0.908±0.017 | 0.251±0.313 |
| R.Amygdala | 0.910±0.017 | 0.197±0.149 | 0.911±0.019 | 0.195±0.197 |
| *Subregions (L,R)* | 0.781±0.069 | 0.722±0.710 | **0.804±0.059** | **0.591±0.622** |
| Lateral | 0.841±0.050 | 0.858±0.712 | **0.862±0.034** | **0.617±0.625** |
| Basal | 0.785±0.059 | 0.365±0.189 | **0.800±0.042** | 0.489±0.692 |
| Cortico-Superficial | 0.728±0.060 | 1.129±0.911 | **0.760±0.049** | 0.889±0.693 |
| Centromedial | 0.771±0.051 | 0.535±0.582 | **0.795±0.058** | 0.370±0.295 |

Table.2. Illustration of the effect of data augmentation on both the amygdala and subnuclear segmentation in terms of DSC and ASSD. Improved values with significance are shown in bold.

*3.3. Comparison with multi-atlas method*

We compared our results after data augmentation with an existing multi-atlas method (Wang et al., 2014). The multi-atlas method was evaluated on our dataset in a leave-one-out scheme. Table 4 shows that our method outperforms the multi-atlas method on both metrics in both amygdala and the subnuclear segmentation. Specifically, on whole amygdala segmentation, our method yielded significantly better mean DSC ($p < 10^{-4}$) and ASSD values ($p < 10^{-4}$) compared to the multi-atlas method. On subnuclear segmentation, our method is also superior to the multi-atlas method on all the four subregions in terms of DSC ($p < 10^{-4}$; $p = 0.001$; $p < 10^{-5}$; $p < 10^{-4}$). The ASSD values of our method are also significantly better than those of multi-atlas method for the basal nuclei and centromedial nuclei ($p < 10^{-3}$; $p = 0.004$). The multi-atlas method yielded slightly better ASSD values on lateral nuclei and cortical-superficial nuclei, while the differences are not significant ($p = 0.412$; $p = 0.682$).

|  | Multi-atlas | | Our Method | |
| --- | --- | --- | --- | --- |
|  | DSC | ASSD/mm | DSC | ASSD/mm |
| *Amygdalae* | 0.881±0.021 | 0.579±0.101 | **0.910±0.018** | **0.223±0.242** |
| L.Amygdala | 0.882±0.019 | 0.571±0.101 | **0.908±0.017** | **0.251±0.313** |
| R.Amygdala | 0.880±0.023 | 0.586±0.104 | **0.911±0.019** | **0.195±0.197** |
| *Subregions (L,R)* | 0.752±0.073 | 0.655±0.182 | **0.804±0.059** | 0.591±0.622 |
| Lateral | 0.803±0.070 | 0.602±0.200 | **0.862±0.034** | 0.617±0.625 |
| Basal | 0.754±0.061 | 0.726±0.162 | **0.800±0.042** | **0.489±0.692** |
| Cortico-Superficial | 0.699±0.057 | 0.751±0.158 | **0.760±0.049** | 0.889±0.693 |
| Centromedial | 0.752±0.064 | 0.541±0.118 | **0.795±0.058** | **0.370±0.295** |

Table.3. Comparison of our method and the multi-atlas method on subnuclear segmentation in terms of DSC and ASSD. Better values with significance are highlighted in bold.

## 3.4. Effects of each pathway

To demonstrate the advantage of our proposed dual-path design, additional training was conducted using each of the pathways separately to investigate the effect of each pathway on the performance. The results were then compared with those of the proposed dual-pathway model. For simplicity, we only analyzed the subnuclear segmentation results across all subjects and did not perform data augmentation. The pathway that has a smaller receptive field was denoted as Pathway_local, and the one that incorporates larger context denoted as Pathway_global. The mean Dice coefficient and average symmetric surface distance (ASSD) of the three models across all four subregions are shown in Fig. 3. Fig. 4 plots the segmentation results generated from the three models for a representative subject. Results at subject level are also discussed, shown in Fig.5.

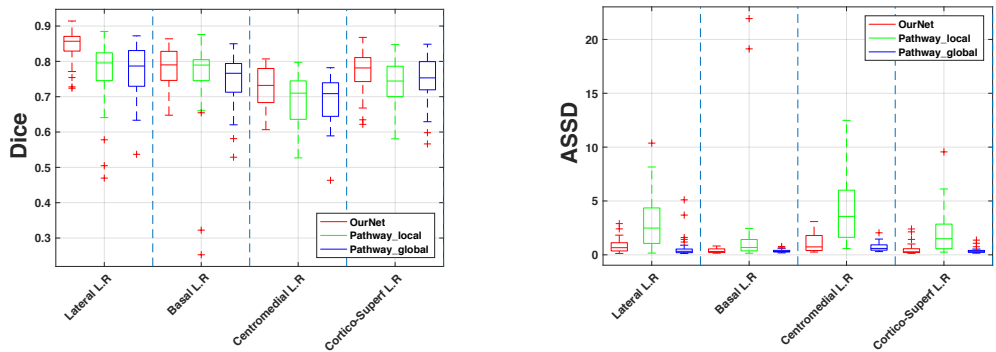

FIg. 3. Evaluation of the effectiveness of each single pathway and the proposed dual-path model. **Structures on the x-axis are listed in descending order by their volume-to-surface ratio.**

Fig.3 shows that pathway_local and pathway_global yielded similar DSC values. In terms of the ASSD, pathway_global outperformed pathway_local on every subregion ($p < 10^{-3}$). With two pathways combined, the DSC values of the proposed dual-path model was superior to each single-path model on all subregions: it outperformed pathway_global on every subregion (p < 0.05), and is significantly better than pathway_local on the lateral, centromedial and cortico-superficial subregions (p < 0.05) while only slightly higher DSC value was obtained on the basal subregion (p = 0.16). The ASSD values of the proposed method were comparable with those of pathway_global (p > 0.05 for all subregions), and were significantly better than those of pathway_local (p < $10^{-3}$ for all subregions). Pathway_local had the worst ASSD values, partially due to the misclassification of distant background voxels, as shown in Fig. 4.

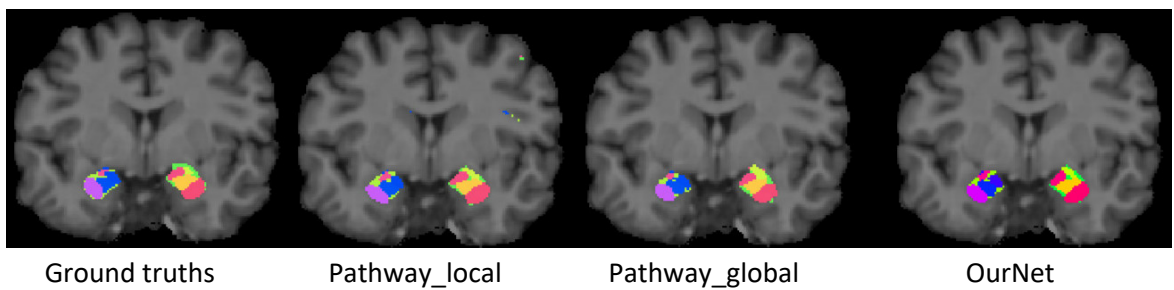

| Ground truths | Pathway_local | Pathway_global | OurNet |

Fig.4. Qualitative segmentation examples generated by each single-path model and the proposed dual-path model, showing influences of each single pathway on the final dual-path model. Specifically, the incorporation of larger context (pathway _global) enabled the final model to better localize the subregions, thus reducing false positives (misclassifications of background voxels), while pathway_local helped capture the appearance details of the final output.

The boxplot in Fig. 5 (a) shows the results of the three models across all subjects. It can be observed that pathway_local yielded higher mean Dice Coefficient than pathway_gobal, though the difference was not significant (p = 0.70). It is noteworthy that two outlier cases scored particularly low for pathway_local. These cases exhibit atypical anatomy close to the amygdala, though not at the amygdala itself. If these two abnormal cases were excluded, pathway_local yields better mean DSC values than Pathway_global (p = 0.156). Meanwhile, it is worth noting that pathway_global and the proposed dual-path model yielded relatively better DSC values on these two abnormal cases compared to pathway_local, as shown in Fig.5 (b), and thus show higher stability than pathway_local.

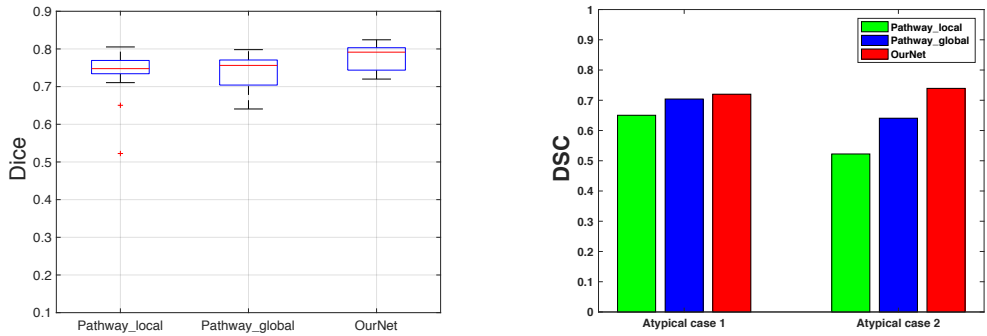

(a) (b)

Fig. 5. (a) Dice values of the two single-path models and the proposed dual-path model in boxplots across all subjects. Outliers are drawn as red '+' symbols; (b) performance of the compared models on two abnormal cases: subject 207 and subject 209.

*3.5. Uncertainty estimation*

As introduced previously, Monte Carlo sampling with dropout is able to provide uncertainty estimation as a visual aid. In particular, we show that this information helped identify atypical cases in training data in our experiment. The uncertainty maps obtained from our network for the two atypical cases mentioned in section 3.4 are shown in Fig.6. The first atypical case has a cyst that causes a significant displacement of the cerebral peduncle that is adjacent to the amygdala. The other atypical case has relative hippocampal head agenesis that changed the shape of the left amygdala as compared to the contralateral one. The uncertainty maps of these two cases show that there is a greater uncertainty to the network and the maps appear noisier than those of typical cases.

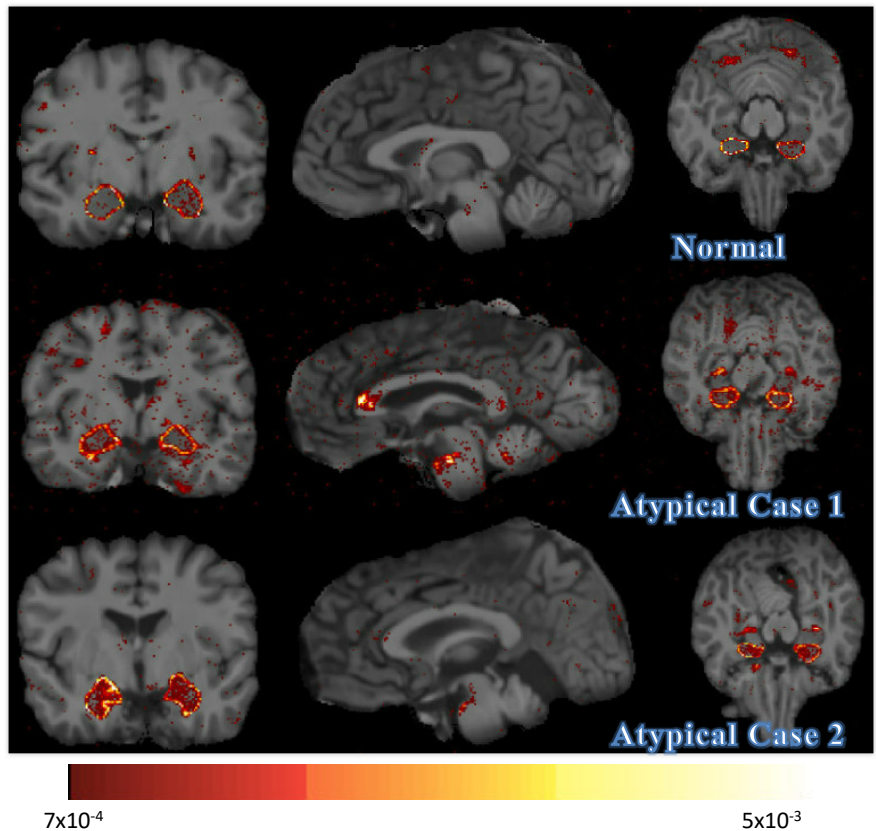

7x10<sup>-4</sup>                                        5x10<sup>-3</sup>

$7\times10^{-4}$       $5\times10^{-3}$

Fig. 6. Illustration of the uncertainty maps of an representative typical case and the two atypical cases.

## 4. Discussion

In this study, we proposed a Bayesian FCNN with dual dilated pathways for the challenging segmentation tasks of the amygdala and its subregions. The proposed model is designed to retain fine details as well as incorporate multi-scale contextual information via oscillating dilated convolutions while alleviating the imbalanced class issue. Additionally, its Bayesian characteristic allows for an uncertainty estimation, which helped identify the atypical cases in our experiments. We evaluated our model on MRI image data acquired from a small cohort of adolescents. The results show that the proposed method demonstrated superior performance as compared to several existing state-of-the-art segmentation methods, as measured by both volumetric and morphological metrics. Our approach is also highly efficient as the GPU based implementation takes less than 1 minute to segment the data of a subject, which is several orders of magnitude faster than multi-atlas based approaches.

The results of using each pathway separately demonstrate that each pathway provides complementary information towards the solution of the segmentation problem, which matches our expectation. Specifically, the pathway_local with a smaller receptive field allows detailed learning of the local appearance but is prone to spatial inconsistencies due to local similarities. The pathway_global that utilizes dilated convolutions to acquire greater context information allows for better localization of the boundary and ROIs but could miss small details. For example, this can be seen from Fig.4 that the challenging residual regions (green) were better recognized by the

pathway_local, while the segmentation results were not spatially consistent; the pathway_global yielded the most spatially consistent results and exhibited better boundary localization, as reflected by its low ASSD values. Combining the appearance and the context information obtained from the two pathways, the proposed model obtained substantially more accurate segmentation results both volumetrically and morphologically. In particular, its ASSD values gained significant improvement over using the pathway_local alone, though it did not exceed the pathway_global. This is likely due to the ASSD sensitivity to the misclassifications of distant small isolation regions. The pathway_global incorporates larger context and thus is especially effective in reducing such false positives (Fig. 4), indicating its strong localization ability for ROIs. We also note that the incorporation of large context may even make network models more robust to atypical cases. This can be seen from the fact that the global context pathway and the dual-pathway model performed a lot better than the local context pathway did on the two atypical cases. Interestingly, we observed similar behaviors between LiviaNET (a single-path model) and the pathway_local, and between DeepMedic (a dual-path model) and the pathway_global, specifically on the second atypical case. We therefore speculate that network models with contextual information may learn more discriminative features about the ROIs and thus become more invariant to nearby changes. It would be interesting to further investigate the influence of the context size on a CNN's performance in atypical cases in the future.

The comparison of the proposed parallel convolutional model and DeepMedic - another dual-path convolutional model also revealed an intriguing phenomenon: the average symmetric surface distance values of our model are better than those of DeepMedic in both tasks, though both approaches incorporate a larger context. This suggests a difference in localization of boundary between down-sampling operation and dilated convolution: down-sampling/max-pooling operation typically discards position information and can only provide coarse low-resolution semantic cues, i.e., the presence of the objects, which makes it hard to delineate the boundaries, as can also be seen in Chen et al. (2015). Dilated convolution, on the other hand, is able to provide precise position cues, in addition to other contextual information such as co-occurrence and scale cues, since it inherently avoids the toll on resolution. For example, by visualizing the segmentation results of a representative subject (Fig.2 (a), (b)), we can see that our method obtained better agreement with contours of the ground truths, while DeepMedic yielded more smooth segmentation results (Fig.2 E).

In comparisons with other deep learning models, our method ranks top among those of Wachinger et al. (2017), Mehta et al. (2017), Kushibar et al. (2018) and Dolz et al. (2017), but it should be noted that the evaluation datasets of most of these methods differ from ours and thus results should be cautiously compared. DeepNAT (Wachinger et al., 2017) consists of three convolutional layers, one max-pooling layer and three fully connected layers, which is quite shallow compared with all the other compared models. This may limit its ability to learn more abstract and complex features (DSC of the Amygdalae estimated from the boxplots = 0.75). M Net (Mehta et al.,2017)'s architecture is inspired by the U-Net, so it also follows an encoder-decoder scheme, and needs to combat with imbalanced class issue using a weighted cross entropy loss function as discussed previously. Its overall Dice Coefficient (DSC = 0.735) on the amygdala is the lowest among all the compared methods, despite its success in segmenting some larger structures, which again confirmed the difficulty in segmenting small structures. Kushibar et al. (2018) employ a multi-path model where each pathway independently processes a slice in a different view. After including an extra path that integrates atlas probabilities for providing spatial priors, its DSC values for the amygdalae gain significant improvement (DSC = 0.83, 0.82), showing the importance of context. LiviaNET (Dolz et al.,2017) is a single-path fully convolutional network. It also preserves resolution by not including any down-sampling operation or strided convolution, which however leads to a small receptive field (19 x 19 x 19). The receptive field of this size may be just sufficient to enclose the whole amygdala, but could cover each subnucleus as a whole as well as model additional context information, which benefits the recognition of the subregions. Moreover, LiviaNET combines fine-grained features from intermediate convolutional layers. It may be all these factors that make it more advantageous in terms of Dice on subnuclear segmentation. However, it is also more prone to spatial inconsistencies because of a small receptive field, as shown in the results (table 2; Fig.3).

The effect of data augmentation was also investigated in our experiments. The results show that data augmentation indeed improved the DSC and ASSD values in both segmentation tasks, especially on the subnuclear segmentation, while the influence on the amygdala segmentation was not substantial. In the future, we plan to delineate the effects of different data augmentation methods (flipping, rotation, scaling etc.) particularly on small structures segmentation, and explore more augmentation methods to further improve the segmentation accuracy.

As a Bayesian FCNN, our model is able to provide voxel-wise uncertainty on each prediction, which can be used for estimating the segmentation quality when ground-truth is not available. Our previous study has investigated the effect of the size of the training set on the uncertainty behavior (Zhao et al., 2018). Here, we show that the uncertainty information helped us identify the abnormal cases in the dataset, which indicates that *high uncertainty does reflects incorrect predictions.* This allows users to decide whether or not to reject the segmentation results with high uncertainty, which greatly facilitates decision-making in biomedical applications where accuracy is critical.

There are several limitations in our study. The biggest limitation is the small sample size of the data which may not capture the variability of other or larger populations. For future training, we will extend our training set, and the segmentation accuracy is likely to be further improved. Another limitation is that we did not carry out a grid search over the dropout rate, so there may exist better-calibrated uncertainty estimates. However, a grid search over the parameters for a large model as ours can be prohibitive as it is extremely time consuming and computational costly. In future work, we plan to investigate the use of the technique proposed in Gal et al.(2017) to allow for automatic tuning of the dropout rate and thus shorten the time for experiments.

## 6. Conclusion
We propose a 3D dual-path Fully-CNN method based on dilated convolutions for segmenting amygdala and its subregions with high accuracy. As a Bayesian FCNN, our network can also produce reliable uncertainty information that can facilitate the evaluation of segmentation of unseen data in the absence of ground-truths, and decision making in broader biomedical applications. Data augmentation has been shown effective particularly on subnuclear segmentation. We believe that the principles of our architecture are not limited to the segmentation of the amygdala and its subregions but could also apply to segmentation of small structures on other digital medical images from macro-to microscopic modalities.

## 7. Acknowledgement

This work was supported by NARSAD: Brain and Behavior grant 24103 (to BN) and National Institutes of Health grant funding NINDS R01 NS092870, NIMH P50 MH100031 and the Waisman Center U54 IDDRC from the Eunice Kennedy Shriver National Institute of Child Health and Human Development (U54 HD090256).